\documentclass[runningheads]{llncs}
\usepackage[T1]{fontenc}
\usepackage{graphicx}
\usepackage{color}
\usepackage{amsmath}
\usepackage{mathtools}
\usepackage{amsfonts}
\usepackage[labelfont=bf]{caption}
\usepackage[labelformat=simple,labelfont=bf]{subcaption}

\usepackage{booktabs,multirow}
\usepackage{array}
\usepackage{pifont}
\usepackage{bbding}
\usepackage[pagebackref=true,breaklinks=true,colorlinks=true,bookmarks=false,citecolor=blue,urlcolor=blue,linkcolor=blue]{hyperref}
\usepackage[misc]{ifsym}
\usepackage{multirow}

\newcommand{\eg}{\textit{e}.\textit{g}.\ }

\begin{document}
\title{Neural LerPlane Representations for \\ Fast 4D Reconstruction of Deformable Tissues} % efficient

\author{
Chen Yang\textsuperscript{1}, 
Kailing Wang\textsuperscript{1},
Yuehao Wang\textsuperscript{2},
Xiaokang Yang\textsuperscript{1},
Wei Shen\textsuperscript{1\dag}
}
\institute{\textsuperscript{1} {MoE Key Lab of Artificial Intelligence, AI Institute, Shanghai Jiao Tong University} \textsuperscript{2} {Dept. of Computer Science and Engineering, The Chinese University of Hong Kong}}

\maketitle              % typeset the header of the contribution
%
%
%
% main body
\begin{abstract}
Reconstructing deformable tissues from endoscopic stereo videos in robotic surgery is crucial for various clinical applications. However, existing methods relying only on implicit representations are computationally expensive and require dozens of hours, which limits further practical applications.
To address this challenge, we introduce LerPlane, a novel method for fast and accurate reconstruction of surgical scenes under a single-viewpoint setting. 
LerPlane treats surgical procedures as 4D volumes and factorizes them into explicit 2D planes of static and dynamic fields, leading to a compact memory footprint and significantly accelerated optimization. 
The efficient factorization is accomplished by fusing features obtained through linear interpolation of each plane and enables using lightweight neural networks to model surgical scenes. Besides, LerPlane shares static fields, significantly reducing the workload of dynamic tissue modeling.
We also propose a novel sample scheme to boost optimization and improve performance in regions with tool occlusion and large motions. Experiments on DaVinci robotic surgery videos demonstrate that LerPlane accelerates optimization by over 100$\times$ while maintaining high quality across various non-rigid deformations, showing significant promise for future intraoperative surgery applications. 
\end{abstract}
\section{Introduction}
% 千万要注意，我们的方法要扣紧1. 医学 2. 加速
% what and why
Reconstructing deformable tissues in surgical scenes accurately and efficiently from endoscope stereo videos is a challenging and active research topic. Such techniques can facilitate constructing virtual surgery environments for surgery robot learning and AR/VR surgery training and provide vivid and specific training for medics on human tissues. Moreover, real-time reconstruction further expands its applications to intraoperative use, allowing surgeons to navigate and precisely control surgical instruments while having a complete view of the surgical scene. This capability could reduce the need for invasive follow-up procedures and address the challenge of operating within a confined field of view.

Neural Radiance Fields (NeRFs)~\cite{mildenhall2021nerf}, a promising approach for 3D reconstruction, have demonstrated strong potential in accurately reconstructing deformable tissues in dynamic surgical scenes from endoscope stereo videos.
EndoNeRF~\cite{endo}, a recent representative approach, represents deformable surgical scenes using a canonical neural radiance field and a time-dependent neural displacement field, achieving impressive reconstruction of deformable tissues. However, the optimization for dynamic NeRFs is computationally intensive, often taking dozens of hours, as each generated pixel requires hundreds of neural network calls. This computational bottleneck significantly constrains the widespread application of these methods in surgical procedures.

% EndoNeRF~\cite{endo}, a recent approach that represents deformable surgical scenes as a canonical neural radiance field along with a time-dependent neural displacement field, has demonstrated impressive performance on dynamic surgical scenes. 
% However, these methods require extensive time and computational resources for optimization, often taking dozens of hours. Since each generated pixel requires more than 100 neural network calls.
% This computational bottleneck limits the wide applicability of these methods.
% However, training and inferring these dynamic fields are time and resource-intensive since 

% Most existing methods model dynamic scenes by introducing an additional deformation network with a similar scale to the radiance network, which maps point coordinates into a canonical space. However, this approach requires significantly more time and resources to train and infer dynamic fields. Moreover, mapping point coordinates into a canonical space are inadequate in representing non-rigid deformations such as distortion and disappearance, which are common in surgical procedures. Consequently, the reconstructed areas have low fidelity.

% claim static accelerate is already in fast developing
Recently, explicit and hybrid methods have been developed for modeling static scenes, achieving significant speedups over NeRF by employing explicit spatial data structures~\cite{hedman2021baking,plenoxels,yu2021plenoctrees} or features decoded by small MLPs~\cite{tensorf,instant-ngp,sun2022direct}. Nevertheless, these methods have only been applied to static scenes thus far.
% claim challenges
Adopting these methods to surgical scenes presents significant challenges for two primary reasons. Firstly, encoding temporal information is essential for modeling surgical scenes while naively adding a temporal dimension to the explicit data structure can significantly increase memory and computational requirements.
Secondly, dynamic surgical scene reconstruction suffers from limited viewpoints, often providing only one view per timestep, as opposed to static scenes, which can fully use multi-view consistency for further regularization. This condition requires sharing information across disjoint timesteps for better reconstruction. 
% \textcolor{red}{add a challenge for tool occlusion ! be more specific on our task! "deformable tissue reconstruction among endoscopy videos"}
% Representing surgical scenes with explicit structures is challenging for two main reasons. Firstly, temporal information is essential for modeling surgical scenes, thus these explicit structures must encode temporal information.
% Naively adding a temporal dimension to the explicit data structure can significantly increase memory and computation requirements. Secondly, dynamic surgical scene reconstruction suffers from limited viewpoints, and treating time steps independently may result in insufficient scene coverage for high-quality reconstruction.

\begin{figure}[t]
\includegraphics[width=\textwidth]{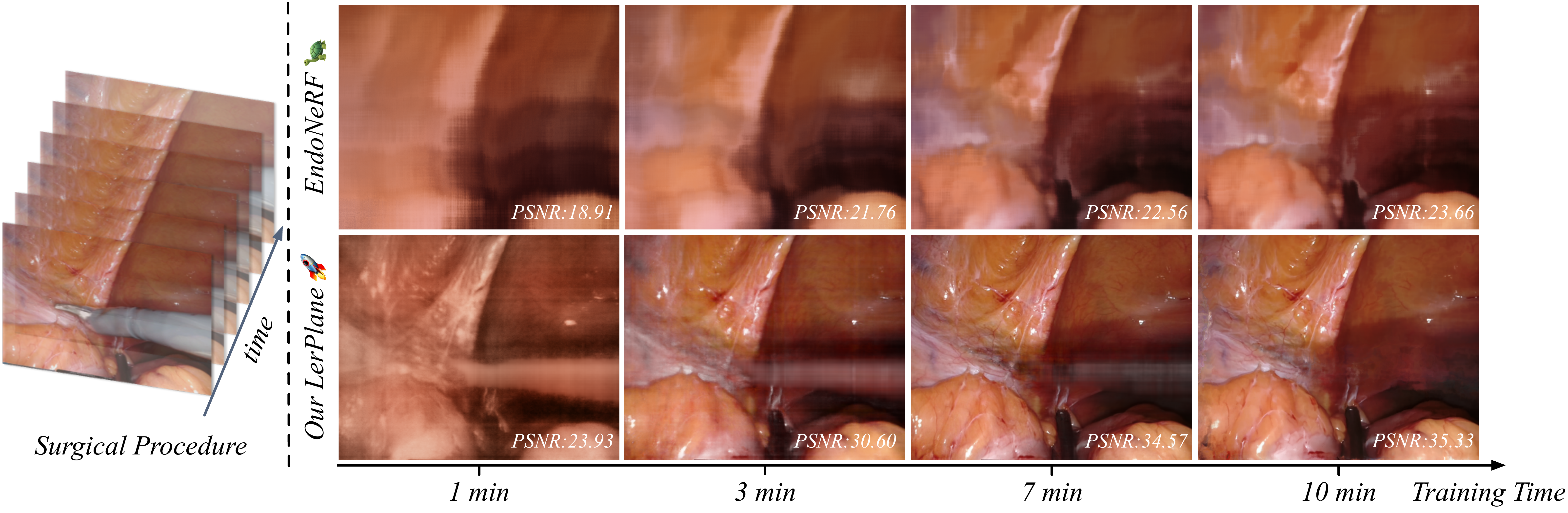}
\caption{Performance along with training time. We show the results of EndoNeRF (top) and LerPlane (bottom) with the same training time. LerPlane exhibits remarkable restoration of surgical scenes with just a few minutes of optimization.} 
\label{fig: teaser}
\vspace{-\baselineskip}
\end{figure}

% claim method
To address the aforementioned challenges, we propose a novel method for fast and accurate reconstruction of deformable tissues in surgical procedures, Neural LerPlane (Linear Interpolation Plane), by leveraging explicitly represented multi-plane fields. 
Specifically, we treat surgical procedures as 4D volumes, where the time axis is orthogonal to 3D spatial coordinates. LerPlane factorizes 4D volumes into 2D planes and uses space planes to form static fields and space-time planes to form dynamic fields. This factorization results in a compact memory footprint and significantly accelerates optimization compared to previous methods~\cite{endo,corona2022mednerf}, which rely on pure MLPs, as shown in Fig.~\ref{fig: teaser}. LerPlane enables information sharing across timesteps within the static field, thereby reducing the negative impact of limited viewpoints.
Moreover, considering the surgical instrument occlusion, we develop a novel sample approach based on tool masks and contents, which assigns higher sampling probability to tissue pixels that have been occluded by tools or have a more extensive motion range. By targeting these regions, our approach allows for more efficient sampling during the training, leading to higher-quality results and faster optimization.

We summarize our contributions: 
\begin{enumerate}
\item A fast deformable tissue reconstruction method, with rendering quality comparable to or better than the previous method in just 3 minutes, which is over 100x faster. 
\item An efficient representation of surgical scenes, which includes static and dynamic fields, enabling fast optimization and high reconstruction quality.
\item A novel sampling method that boosts optimization and improves the rendering quality.
Compared to previous methods, our LerPlane, achieves much faster optimization with superior quantitative and qualitative performance on 3D reconstruction and deformation tracking of surgical scenes, providing significant promise for further applications.
\end{enumerate}
% \begin{enumerate}
%     \item We significantly accelerate the optimization period of surgical scene reconstruction with pure python implementation, where our LerPlane achieves better or similar rendering quality by taking only 10 minutes, achieving 100x faster than EndoNeRF. 
%     \item We propose a novel representation of surgery scenes, consisting of static and dynamic fields, providing fast optimization, high reconstruction quality, and strong interpretability.
%     \item We propose a spatiotemporal importance sampling to boost the optimization and improve the rendering quality among regions that are occluded by tools and enable large motion.
% \end{enumerate}
\section{Method}
\subsection{Overview} \label{sec: 2.1 overview}
\begin{figure}[t]
\includegraphics[width=0.9\textwidth]{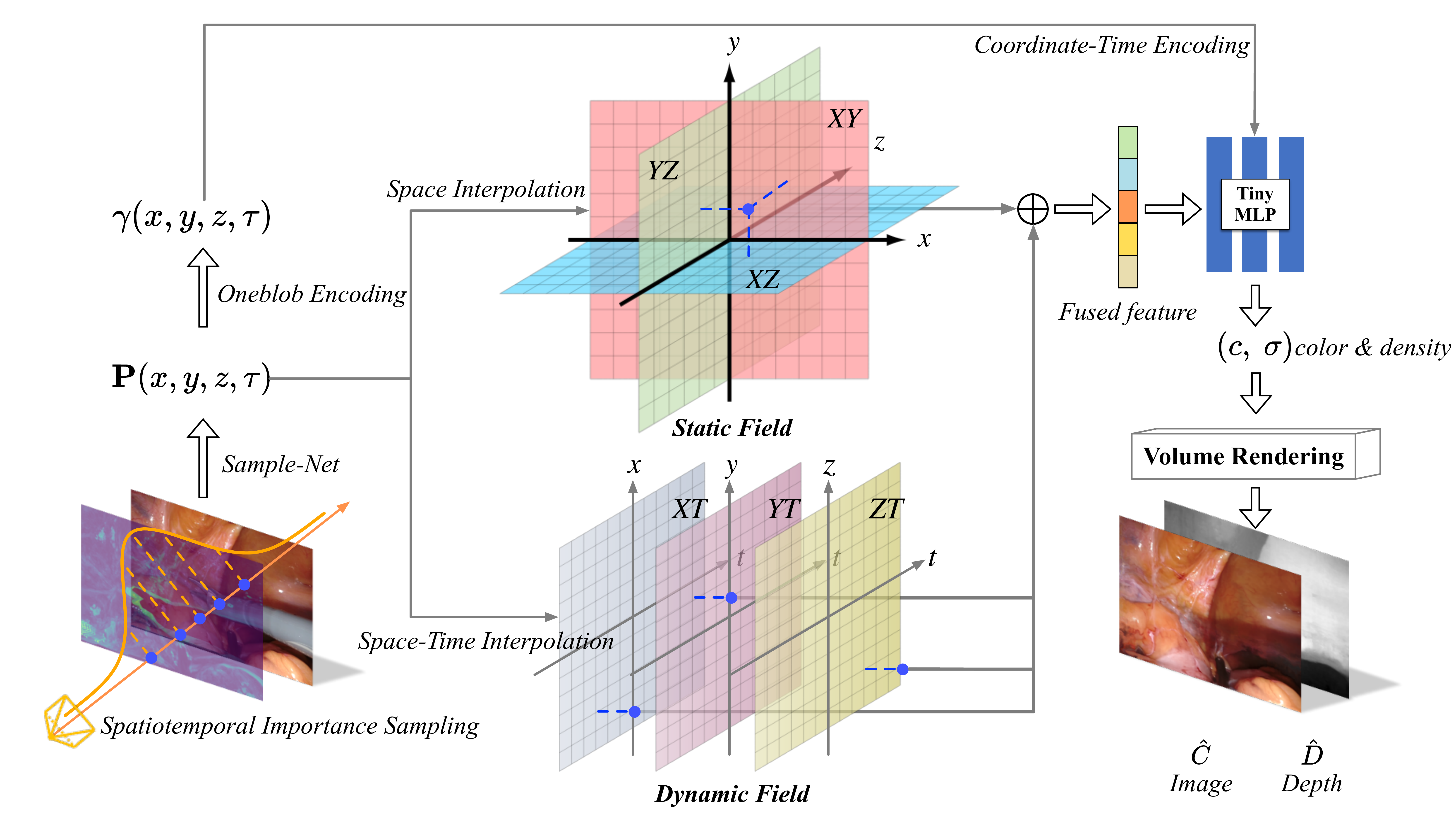}
\caption{Illustration of our fast 4D reconstruction method, LerPlane.
} \label{fig: overview}
\vspace{-\baselineskip}
\end{figure}

LerPlane represents surgical procedures using static and dynamic fields, each of which is made up of three orthogonal planes (Sec.~\ref{sec: 2.3 Neural LerPlane Representations for Deformable Tissues}). 
It starts by using spatiotemporal importance sampling to identify high-priority tissue pixels and build corresponding rays (Sec.\ref{sec: 2.4 Spatiotemporal Importance Sampling}). Then we sample points along each ray and query features using linear interpolation to construct fused features. The fused features and encoded coordinate-time information are input to a lightweight MLP, which predicts color and density for each point (Sec.\ref{sec: 2.5 Space-Time Positional Encoding}). 
To better optimize LerPlane, we introduce some training schemes, including sample-net, various regularizers, and a warm-up training strategy (Sec.\ref{sec: 2.6 Optimization}). Finally, we apply volume rendering to produce predicted color and depth values for each chosen ray.
The overall framework is illustrated in Fig.~\ref{fig: overview}.

\subsection{Preliminaries} \label{sec: 2.2 Preliminaries}
% volume rendering
Neural Radiance Field (NeRF)~\cite{mildenhall2021nerf} is a coordinate-based neural scene representation optimized through a differentiable rendering loss. 
NeRF maps the 3D coordinate and view direction of each point in the space into its color values $\boldsymbol{c}$ and volume density $\sigma$ via neural networks $\Phi_r$. 
\begin{equation}
    \boldsymbol{c}, \sigma=\Phi_{r}(x, y, z, \theta, \phi). \label{eq:nerf_pred}
\end{equation}
It calculates the expected color $\hat{C}(\mathbf{r})$ and the expected depth $\hat{D}(\mathbf{r})$ of a pixel in an image captured by a camera by tracing a ray $\mathbf{r}(t) = \mathbf{o} + t\mathbf{d}$ from the camera center to the pixel. Here, $\mathbf{o}$ is the ray origin, $\mathbf{d}$ is the ray direction, and $t$ is the distance from a point to the $\mathbf{o}$, ranging from a pre-defined near bound $t_n$ to a far bound $t_f$. $w(t)$ represents a weight function that accounts for absorption and scattering during the propagation of light rays. The pixel color is obtained by classical volume rendering techniques~\cite{kajiya1984ray}, which involve sampling a series of points along the ray.
\begin{equation}
    \hat{C}(\mathbf r) = \int_{t_{n}}^{t_{f}}\!  w(t) \mathbf{c}(t) d t, 
    \hat{D}(\mathbf r) = \int_{t_{n}}^{t_{f}}\!  w(t) t d t,
    w(t)=\exp \left(-\int_{t_{n}}^{t}\!  \sigma(s) d s\right) \sigma(t). \label{eq:volume_render}
\end{equation}
% NeRF connects real-world 3D points with image pixels by accumulating colors $\boldsymbol{c}_{i}$ and densities $\sigma_{i}$ of sample points along the ray.

\subsection{Neural LerPlane Representations for Deformable Tissues} \label{sec: 2.3 Neural LerPlane Representations for Deformable Tissues}
% We represent a surgical scene using six orthogonal feature planes: three space planes (\textit{i.e.} \textit{XY}, \textit{YZ}, \textit{XZ}) are used to construct the static field, and three space-time planes (\textit{i.e.} \textit{XT}, \textit{YT}, \textit{ZT}) are adopted to construct the dynamic field. Each feature plane has a shape of $N \times M \times D$, where $N$ and $M$ represent spatial and temporal resolution, respectively, and $D$ is the size of the feature. 
% Given one pixel $p_{ij}$ with color $C$ in specific timestep $\tau$, we first cast a ray $\mathbf{r}(t) = \mathbf{o} + t\mathbf{d}$  from camera origin to the pixel. Then we sample spatial-temporal points along the ray with its 4D coordinates. LerPlane computes a feature vector for a point $\mathbf{P} (x, y, z, \tau)$ by projecting the point onto each feature plane and using bilinear interpolation $\mathcal{B}$ to query features among the six feature planes.
A surgical procedure can be represented as a 4D volume, and we factorize the volume into 2D planes.
Specifically, We represent a surgical scene using six orthogonal feature planes, consisting of three space planes (\textit{i.e.}, \textit{XY}, \textit{YZ}, and \textit{XZ}) for the static field and three space-time planes (\textit{i.e.}, \textit{XT}, \textit{YT}, and \textit{ZT}) for the dynamic field. 
Each space plane has a shape of $N \times N \times D$, and each space-time plane owns a shape of $N \times M \times D$, where $N$ and $M$ represent spatial and temporal resolution, respectively, and $D$ is the size of the feature.

To extract features from an image pixel $p_{ij}$ with color $C$ at a specific timestep $\tau$, we first cast a ray $\mathbf{r}(t)$ from $o$ to the pixel. We then sample spatial-temporal points along the ray, obtaining their 4D coordinates. We acquire a feature vector for a point $\mathbf{P}(x, y, z, \tau)$ by projecting it onto each plane and using bilinear interpolation $\mathcal{B}$ to query features from the six feature planes.
\begin{equation}
    \mathbf{v}(x,y,z,\tau) = \mathcal{B}(F_{\textit{XY}}, x, y) \odot \mathcal{B}(F_{\textit{YZ}}, y, z) \ldots \mathcal{B}(F_{\textit{YT}}, y, \tau) \odot \mathcal{B}(F_{\textit{ZT}}, z, \tau), \label{interpolate}
\end{equation}
where the $\odot$ represents element-wise multiplication, inspired by~\cite{chan2022efficient,peng2020convolutional}. The fused feature vector $\mathbf{v}$ is then passed to a tiny MLP $\Theta$, which predicts the color $c$ and density $\sigma$ of the point. Finally, we leverage the Eq.~\ref{eq:volume_render} to get the predicted color $\hat{C}(\mathbf{r})$. Inspired by~\cite{instant-ngp,liu2020neural}, we build the feature planes with multi-resolution planes, \eg $F_{XY}$ is represented by planes with $N = 128$ and $256$.

% claim why this is effective
Existing methods~\cite{endo,corona2022mednerf} for reconstructing surgical procedures using pure implicit representations, requires traversing all possible positions in space-time, which is highly computationally and time-intensive. In contrast, LerPlane decomposes the surgical scene into six explicitly posed planes, resulting in a significant reduction in complexity and a much more efficient representation. This reduces the computational cost from $O(N^3)$ to $O(N^2)$ and enables the use of smaller neural networks, leading to a considerable acceleration in the training period.
% Previous methods, such as \cite{endo,corona2022mednerf}, have shown impressive results in reconstructing surgical procedures using pure implicit representations. However, they require traversing all possible positions in space-time, which is highly computationally and time-intensive.
% In contrast, LerPlane decomposes the surgical scene into six explicitly posed planes. This significant reduction in complexity allows for a much more efficient representation, reducing the computational cost from nearly infinite to $O(N^2)$. As a result, LerPlane can utilize much smaller neural networks, leading to a considerable acceleration in the training period.
% Previous methods, such as \cite{endo,corona2022mednerf}, have shown impressive results in reconstructing surgical scenes using pure implicit representations. 
% To achieve proper convergence, these methods must traverse all possible positions in space and time, which is highly computationally and time-intensive.
% In contrast, our LerPlane decomposes the surgical scene into six explicitly posed planes, significantly reducing the computational cost of modeling the surgical procedures, from nearly infinite to $O(N^2)$. This advantage enables LerPlane to use smaller MLPs than those used in EndoNeRF, leading to an enormous acceleration in the training period.
Besides, methods~\cite{endo,tineuvox,dnerf,dynerf,park2021hypernerf,park2021nerfies} using a single displacement field to supplement the static field struggle with handling variations in scene topology, such as non-linear deformations. In contrast, LerPlane can naturally model these situations using a dynamic field.
\subsection{Spatiotemporal Importance Sampling} \label{sec: 2.4 Spatiotemporal Importance Sampling}
% We propose a novel sampling method to address the tool occlusion and further accelerate the optimization. 
% There are many pixels of surgical tools among endoscopy videos, which is not expected for further observation and cannot support surgery robot learning. Thus we leverage binary masks $\left\{\boldsymbol{M}_{i}\right\}_{i=1}^{T}$ where 1 stands for tissue pixels and 0 stands for tool pixels to reduce influence of tools. Besides, inspired by~\cite{dynerf}, we propose to leverage the temporal difference among frames to further accelerate the optimizing period. Specifically, we create sampling weight maps $\{\mathbf{W}\}_{i=1}^T$, where every element represents the probability of the corresponding pixel/ray be choosen.
% Tool occlusion often occurs in robotic surgery, the number of times the occluded tissue appears in the training set is far less than that of the unobstructed area, which poses a challenge for reconstructing these tissues. 
% We also observe that there are plenty of tissues remain stationary temporally, repeated training of these pixels cannot provide gain for convergence，which slows down the convergence speed.
% In robotic surgery, tool occlusion is a common occurrence that poses a challenge for reconstructing occluded tissues since these tissues appears far less frequently in the training set than others, which means the different pixels have different learning difficulties.
% in the unobstructed area.
Tool occlusion in robotic surgery poses a challenge for reconstructing occluded tissues due to their infrequent occurrence in the training set, resulting in varied learning difficulties for different pixels.
Besides, we observe that many tissues remain stationary over time, and therefore repeated training on these pixels contributes minor to the convergence, reducing efficiency. We design a novel spatiotemporal importance sampling strategy to address the issues above. 
% \textcolor{blue}{In surgical scenes, the existence of tools occludes certain tissue, thereby affecting observations. However, we observe that the tissue typically remains stationary during tool movement, and thus it should be possible to reconstruct the occluded parts utilizing the image on other timestamps at which these parts are not occluded. Also, since stationary parts are easier to optimize compared to dynamic parts, we focus more on the moving parts of the tissue in hope of improve the efficiency overall. To enhance the reconstruction of the parts temporarily covered by the tool and accelerate optimization, }
% \textcolor{red}{Todo, please make the motivation more clear!!!} avoid the inclusion of surgical tool pixels, where 1 indicates tissue pixels and 0 indicates tool pixels. 
In particular, we utilize binary masks $\{\boldsymbol{M}_i\}_{i=1}^T$ and temporal differences among frames to generate sampling weight maps $\{\mathbf{W}\}_{i=1}^T$. These weight maps represent the sampling probabilities for each pixel/ray, drawing inspiration from~\cite{endo,dynerf}.
% utilize binary masks $\{\boldsymbol{M}_i\}_{i=1}^T$ and temporal differences among frames to get sampling weight maps $\{\mathbf{W}\}_{i=1}^T$ which represent sampling probabilities for each pixel/ray, inspired by~\cite{endo,dynerf}.
% Inspired by~\cite{dynerf}, we utilize the temporal differences among frames to further accelerate optimization. In particular, we leverage sampling weight maps $\{\mathbf{W}\}_{i=1}^T$ to assign probabilities to each pixel/ray for selection during training. 
One sampling weight map $\mathbf{W}_i$ can be determined by:
\begin{equation}
    \mathbf{W}_i = \min(\max_{\substack{i-n<j\\<i+n}} (\|\ \mathbf{I}_i\boldsymbol{M}_i - \mathbf{I}_j\boldsymbol{M}_j \|_1)/3, \alpha)\cdot\mathbf{\Omega}_i,~\mathbf{\Omega}_i = \beta(\boldsymbol{M}_iT/ {\sum\limits_{i=1}^T\boldsymbol{M}_i}),
\end{equation}
where $\alpha$ is a lower-bound to avoid zero weight among unchanged pixels, $\mathbf{\Omega}_i$ specifies higher importance scaling for those tissue areas with higher occlusion frequencies, and $\beta$ is a hyper-parameter for balancing augmentation among frequently occluded areas and time-variant areas. 
By unitizing spatiotemporal importance sampling, LerPlane concentrates on tissue areas and speeds up training, improving the rendering quality of occluded areas and prioritizing tissue areas with higher occlusion frequencies and temporal variability.
% we reduce the impact of tool occlusion and reduce the training time on still pixels, so as to improve the convergence speed and improve the rendering quality of occluded areas.
% By assigning probabilities to each pixel/ray using sampling weight maps, we ensure that tissue areas with higher occlusion frequencies and more time-variant than others are more likely to be sampled during training. 
% This way, we not only ..., but also ..., achieving ...
% and $\beta$ are hyperparemters for 
% our design tends to sample those pixels that are more timevariant than others.
% by multipy $1 - \left\{\boldsymbol{M}_{i}\right\}_{i=1}^{T}$ and 
% Thus we propose Spatiotemporal Importance Sampling to address the tool occlusion and further accelerate the optimization speed. Given images $\{\mathbf{I}_i\}_{i=1}^T$ with its corresponding binary masks $\left\{\boldsymbol{M}_{i}\right\}_{i=1}^{T}$ where 0 stands for tissue pixels and 1 stands for tool pixels, we first 
% while our goal is to reconstruct underlying tissues. Thus, training on these tool pixels is unexpected. Our main idea for solving this issue is to bypass those rays traveling through tool pixels over the training stage.
\subsection{Coordinate-Time Encoding} \label{sec: 2.5 Space-Time Positional Encoding}
% We have abandoned the view-direction-based encoding utilized in~\cite{endo} and instead propose Coordinate-Time Encoding in hope of achieving better performance during training. In typical surgery scenes, the camera's perspective should not undergo significant changes. Encoding view direction will redundant structures in the network or even cause the network to learn irrelevant information. By encoding coordinates and time, the network can learn based on the coordinate information for the stationary part of the scene while learning based on time information for the dynamic part of the scene. Moreover, our encoding method can help the network converge faster compared with view-direction-based encoding. 
Previous methods~\cite{endo,corona2022mednerf} apply positional encoded view direction $\gamma(\mathbf{d})$ to model view-dependent appearance. However, during endoscopic operations, camera movements are restricted. The view direction changes are typically minimal. Instead of view encoding, we propose using $\gamma(x,y,z,\tau)$ to enhance the spatiotemporal information.
% \textcolor{blue}{In typical surgical scenes, there are often very slight view direction changes. The view-direction-based encoding used in previous methods has no significant effect, and may even introduce redundant or irrelevant information into the network. To address this issue, we we propose a new method that disables view-direction-based encoding.\\
% The LerPlane has both static field and dynamic field, which inspires us to encode the coordinate information to train the static field and the time information to train the dynamic filed.  To accomplish this, we employ one-blob encoding~\cite{one-blob} for both the coordinate and time dimensions. Our approach encodes more relevant information than view-direction-based encoding, leading to faster convergence.}
% We propose a coordinate-time encoding scheme to replace the view-direction-based encoding used in previous methods, which may lead the network to learn redundant or irrelevant information, particularly in typical surgical scenes with minimal view direction changes.
% To address this issue, we disable the view-direction-based encoding in our proposed method and instead encode the coordinates and time dimensions using one-blob encoding~\cite{one-blob}.
Specifically, the encoding along with the fused features $\mathbf{v}$ from feature planes is input to the MLP $\Theta$, which predicts $\sigma$ and $c$ of each point. Then we utilize Eq.~\ref{eq:volume_render} to render the expected color $\hat{C}$ and depth $\hat{D}$ of one specific ray.

% The encoding enables LerPlane to learn based on coordinate information for the stationary part of the scene and time information for the dynamic part and achieves a faster convergence compared to view-direction-based encoding.

% In our work, we utilized both frequency encoding and one-blob encoding. Function \ref{freq_encode} illustrates the frequency encoding method commonly employed in traditional NeRF.
% \begin{equation}
%     \gamma(p)=\left(\sin \left(2^0 \pi p\right), \cos \left(2^0 \pi p\right), \cdots, \sin \left(2^{L-1} \pi p\right), \cos \left(2^{L-1} \pi p\right)\right)
%     \label{freq_encode}
% \end{equation}
%  one-blob encoding is a recently proposed encoding method that utilizes a single blob of features to represent each vector in the scene. one-blob encoding is believed to have a more accurate fit while not suffering from stripe artifacts like Frequency encoding. Function \ref{one-blob_encode} describes the method explicitly.
% \begin{equation}
% \mathbf{b}(\mathbf{x}) = \frac{1}{|\Omega(\mathbf{x})|} \sum_{\mathbf{y} \in \Omega(\mathbf{x})} \mathbf{f}(\mathbf{y}),
% \label{one-blob_encode}
% \end{equation}
% where $\mathbf{x}$ is the vector to be encoded, $\Omega(\mathbf{x})$ is a local region surrounding the vector, $|\Omega(\mathbf{x})|$ is the number of vectors in the region, and $\mathbf{f}(\mathbf{y})$ is a vector of features extracted from a vector $\mathbf{y}$ in the region.
\subsection{Optimization} \label{sec: 2.6 Optimization}
We adopt a joint supervision approach to optimize the tiny MLP $\Theta$ and feature planes using rendered color and depth. To further improve the optimization process, we propose several optimization schemes, including a sample-net for better-sampled points, a warm-up strategy to address outliers, and several regularizers.
% We optimize the $\Theta$ and features planes by jointly supervising the rendered color and depth. To effectively train LerPlane, we introduce some optimization schemes, which include a sample-net for better-sampled points, a warm-up strategy to address outliers, and several regularizers.

\textit{Sample-Net.}
The sampling of spatiotemporal points is crucial for volume rendering, with a particular focus on sampling around tissue regions for optimal performance. We replaced the conventional two-stage time-consuming sampling strategy with a single sample-net and train it using histogram loss~\cite{mip360}. The sample-net is a lightweight single-resolution LerPlane model that provides more accurate sampling points for the full model.
% \textcolor{blue}{We aim to improve the quality of the sampling points during training, as the details of the surgical scene are concentrated on tissue surfaces. To achieve this, }
% we propose an efficient NeRF-based surgical scene reconstruction approach by using a small-resolution LerPlane model as a sample-net, which is trained with the histogram loss~\cite{mip360} and a depth loss. Instead of the conventional two-step approach, sample-net adaptively passes samples from initial uniform sampling points to the multi-resolution full model.
% This results in sharper details by placing samples closer to tissue surfaces

\textit{Regularizers.}
We apply some regularization to address the limited information available in surgical scene reconstruction. We adopt 2D total variation (TV) loss for space planes in~\cite{plenoxels,tensorf,VoxGRAF} and 1D TV loss on the space axis for space-time planes and a similar smooth loss on the time axis. Additionally, we introduce a minor time-invariant loss to separate the static and dynamic fields as much as possible, encouraging the features in space-time planes to remain unchanged.

\textit{Warm-up Training Strategy.}
Since single-view captures cannot provide valid scale information, we leverage pseudo ground truth depth maps $D(\mathbf{r})$ generated by STTR-light~\cite{sttrlight} from stereo images to guide the optimization. Specifically, we apply a Huber loss for depth regularization:
\begin{equation}
    \mathcal{L}_{D}=\left\{\begin{array}{ll}
0.5\Delta D(\mathbf{r})^{2}, & \text { if }\left|\Delta D(\mathbf{r})\right|<\delta  \\
\delta  \cdot\left(\Delta D(\mathbf{r})-0.5 \cdot\delta \right), & \text { otherwise }
\end{array}\right.
\end{equation}
where $\Delta D(\mathbf{r})=|\hat{D}(\mathbf{r})-D(\mathbf{r})|$ represents the absolute depth difference among valid depth values, $\delta$ is a threshold at which to change loss type.
Considering that the predicted depth maps encounter a lot of unreliable depth values and missing areas~\cite{endo}, we design a simple by effective warm-up training strategy. 
Specifically, we apply the $\mathcal{L}_{D}$ to depths from both the sample-net and the full model during the first half of the training. In the remaining iterations, we disable the $\mathcal{L}_D$ and use other regularization to refine unreliable depths.

% During the first half of the training, the $\mathcal{L}_{D}$ is applied on depths from both the sample-net and the full model during the first half of the training. In the remaining iterations, we disable the depth regularization, refining the unreliable depths via other regularizations. 

% Speciﬁcally, we introduce the sparse depth loss item L sparse only in the ﬁrst half of the training ( λ sparse = 0.05). In the remaining iterations, we set λ sparse = 0 and let L ph and L pc reﬁne the depths of pixels where noisy points are located.

% As mentioned in Sec.~\ref{sec: 2.2 Preliminaries}, the emitted color $\hat{C}(\mathbf{r})$ and optical depth $\hat{D}(\mathbf{r})$ of one specific ray can be computed by volume rendering. Thus we leverage STTR-light~\cite{sttrlight} to generate pseudo ground truth depth maps $D(\mathbf{r})$ from stereo images
% serving as ground truth depth map to supervise the scene reconstruction and address the scale problem from single-view captures.

% LerPlane leverages estimated stereo depth maps from STTR-light~\cite{sttrlight} to address the scale problem in monocular case. Note that the 

% We represent dynamic 3D scenes using the proposed HexPlane, which is optimized by photometric loss between rendered and target images. For point (x, y, z, t), its opacity and appearance feature are quired from HexPlane, and the ﬁnal RGB color is regressed from a tiny MLP with appearance feature and view direction as inputs. With points’ opacities and colors, images are rendered via volumetric rendering. The optimization objective is:

\section{Experiments}
\subsection{Dataset and Evaluation Metrics}
% data
We evaluate our proposed method on the EndoNeRF dataset~\cite{endo}, a collection of typical robotic surgery stereo videos captured from stereo cameras at a single viewpoint during in-house DaVinci robotic prostatectomy procedures, which is designed to capture challenging surgical scenes with non-rigid deformation and tool occlusion. 
% The dataset comprises six clips with 807 frames, each lasting 4-8 seconds at a frame rate of 15 frames per second.
We evaluate our proposed method by comparing it to existing methods~\cite{edssr,endo} using standard image quality metrics following~\cite{endo}, including PSNR, SSIM, and LPIPS. Additionally, to measure the consistency of the underlying 3D scene, we supplement these metrics using the FLIP metric~\cite{andersson2020flip,andersson2021visualizing}. 
% FLIP measures the number of pixels that change in the image when viewed from the opposite side of the scene, providing a more comprehensive evaluation of 3D reconstruction. 
For qualitative evaluation, we follow the exhibition method from~\cite{endo}.

\begin{figure}[t]
\includegraphics[width=0.9\textwidth]{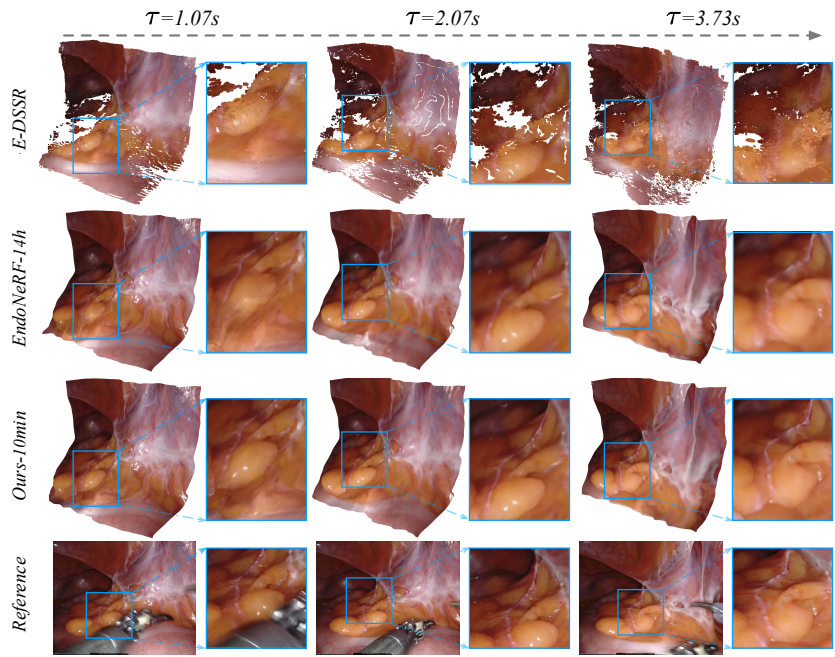}
\caption{Qualitative results on scene ``traction'' from different timesteps $\tau$. } 
\label{fig: qualitative}
\vspace{-\baselineskip}
\end{figure}

\subsection{Implementation Details}
We normalize the scene into device coordinates (NDC) to handle single-view endoscopy videos and then project rays within the NDC space. 
The video duration is normalized to $[-1, 1]$. We use a two-stage sampling network with $128$ and $256$ dimensional plane features for the sample-net. Oneblob encoding~\cite{one-blob} is applied to encode the spatiotemporal information. The full model consists of four resolutions, $64$, $128$, $256$, and $512$ dimensions among space. Hyperparameters include $D=32$ for feature planes, $j=25$ for spatiotemporal importance sampling, $\xi=16$ for Oneblob encoding dimensionality, and $\delta = 0.2$ for depth loss across all experiments. 
An Adam~\cite{adam} optimizer is used with default values for optimization. In each iteration, $2048$ rays are randomly sampled from the whole dataset to form a batch. 
The initial learning rate is set to 0.01. We apply a cosine schedule with a $512$ iterations warming-up stage.
We train all scenes with $9k$ and $32k$ iterations, which take around $3$ and $10$ minutes, respectively, on a single RTX 3090 GPU running the Ubuntu 20.04. Our LerPlane is implemented with pure Pytorch~\cite{Paszke_PyTorch_An_Imperative_2019}. The code is available at \url{https://github.com/Loping151/LerPlane}.
% For further detail on loss weights, please refer to our supplementary materials.

\begin{table}[t]
    \setlength\tabcolsep{5pt}
    \centering
    \caption{Quantitative results on the EndoNeRF dataset. Please refer to Sec.~\ref{sec: 3.4} for explanations of the acronyms used.}
    \resizebox{\linewidth}{!}{{\begin{tabular}{c|cccc|c}
\hline
Methods  & PSNR$\uparrow $        & SSIM$\uparrow$         & LPIPS$\downarrow$         & FLIP$\downarrow$            & Time$\downarrow$            \\ \hline
E-DSSR\cite{edssr}   & 13.398{\tiny $\pm$}1.387   & 0.630 {\tiny $\pm$}0.057  & 0.423 {\tiny $\pm$}0.047  & \textbackslash{} & \textbackslash{} \\
EndoNeRF\cite{endo} & 29.272{\tiny $\pm$}2.836   & 0.921 {\tiny $\pm$}0.022  & 0.088 {\tiny $\pm$}0.020  & 0.085{\tiny $\pm$}0.018    & 14 $h$             \\ \hline
Ours-NS  & 31.532{\tiny $\pm$}1.665   & 0.886{\tiny $\pm$}0.021   & 0.142 {\tiny $\pm$}0.020  & 0.112{\tiny $\pm$}0.016      & 3 $min$            \\
Ours-TS  & 31.544{\tiny $\pm$}1.669   & 0.886{\tiny $\pm$}0.021   & 0.142{\tiny $\pm$}0.020   & 0.111{\tiny $\pm$}0.015      & 3 $min$            \\
Ours-VE  & 32.353{\tiny $\pm$}1.742   & 0.897{\tiny $\pm$}0.022   & 0.131{\tiny $\pm$}0.024   & 0.103{\tiny $\pm$}0.012      & 3 $min$            \\
Ours-NE  & 32.230{\tiny $\pm$}1.655   & 0.895{\tiny $\pm$}0.023   & 0.131{\tiny $\pm$}0.020   & 0.102{\tiny $\pm$}0.012      & 3 $min$            \\
% Ours-NW  & 31.908{\tiny $\pm$}1.521   & 0.889{\tiny $\pm$}0.023   & 0.137{\tiny $\pm$}0.027   & 0.108{\tiny $\pm$}0.013      & 3 $min$            \\ 
\hline
Ours-9k  & 32.589{\tiny $\pm$}1.451   & 0.901{\tiny $\pm$}0.021  & 0.126{\tiny $\pm$}0.028    & 0.103{\tiny $\pm$}0.014      & 3 $min$            \\
Ours-32k & 35.504{\tiny $\pm$}3.076   & 0.935{\tiny $\pm$}0.026  & 0.083{\tiny $\pm$}0.022    & 0.075{\tiny $\pm$}0.031    & 10 $min$           \\ \hline
\end{tabular}
}} %\scriptsize 
    \label{tab:ablation}
    \vspace{-\baselineskip}
\end{table}

\subsection{Evaluation}
We compare our proposed method, LerPlane, against two existing SOTA methods: the surfel warping-based method, E-DSSR~\cite{edssr} and NeRF-based method EndoNeRF~\cite{endo}. We find that E-DSSR struggles to completely reconstruct surgical scenes, resulting in many holes and noisy points (see Fig.~\ref{fig: qualitative}), which leads to poor numerical performance. In contrast, EndoNeRF achieves high-fidelity reconstruction of deformable tissues but requires around 14 hours of optimization, which is computationally expensive and constrains intraoperative use.
LerPlane, on the other hand, achieves comparable results to EndoNeRF with only 3 minutes of optimization, providing nearly 280-fold acceleration. Moreover, with a longer optimization time of 10 minutes, LerPlane outperforms both E-DSSR and EndoNeRF in terms of all metrics, as shown in Table~\ref{tab:ablation}. Our novel importance sampling and encoding strategies further enhance the ability of LerPlane to preserve details and produce accurate visualizations of deformable tissues, as demonstrated in Fig.~\ref{fig: qualitative}.
Our results demonstrate that LerPlane achieves significantly faster optimization without compromising reconstruction quality, showing great potential for future clinical applications in robotic surgery.

\subsection{Ablation Study}
\label{sec: 3.4}
We conduct ablation studies on the EndoNeRF dataset to understand the key components and demonstrate their effectiveness. Table~\ref{tab:ablation} shows the performance of all experiments.
\begin{enumerate}
\item \textit{Sampling Strategy.} We compare with two different methods: naively avoiding tool masks, assigning equal weights to other pixels (Ours-NS), and assigning higher probabilities to highly occluded areas (Ours-TS), as in ~\cite{endo}. Our method effectively prioritizes time-variant and highly occluded areas, significantly improving convergence speed.
% We conduct an ablation study on the sampling strategy among pixels, considering three different approaches, \ie naively avoiding all tool masks, assigning equal weights to other pixels (Ours-NS), and assigning higher sampling probability to areas with higher occlusion frequencies, as proposed in~\cite{endo} (Ours-TS) and our sampling. Experiments demonstrate that our sampling method effectively prioritizes time-variant and highly occluded areas, significantly improving convergence speed.
\item \textit{Encoding Strategy.} Experiments showed that coordinate-time encoding achieves better performance in all metrics compared to no encoding (Ours-NE) or direction encoding (Ours-VE), showing the effectiveness of the proposed encoding. 
\end{enumerate}
% For further detail on loss weights, please refer to our supplementary materials.
Further analysis of the optimization schemes is available in the Supplementary Materials.
% We study the effectiveness of the proposed encoding. Experiments with the direction encoding (Ours-VE) and without any encoding (Ours-NE) achieve similar performance. In contrast, ours-full achieves better performance in terms of all metrics. These results validate the effectiveness of the proposed coordinate-time encoding and support our insight that encoding the coordinates and time dimensions is more appropriate for representing surgical procedures.
% 3) \textit{Warm-up Training Strategy.} We demonstrate that our warm-up training strategy improves reconstruction performance and avoids incorrect depth values, as seen in Table~\ref{tab:ablation} (Ours-NW).
% We show the effectiveness of our warm-up training strategy in Table.~\ref{tab:ablation}. Without warm-up (Ours-NW), incorrect depth values may degrade reconstruction performance.

% \subsection{Extension}
% split static and dynamic.

\section{Conclusion and Future Work}
% In this paper, we present LerPlane, a novel method for fast and accurate reconstruction of surgical scenes from single-viewpoint endoscopic stereo videos. By leveraging explicitly represented multi-plane fields, we factorize 4D volumes into static and dynamic fields, significantly accelerating optimization while maintaining high reconstruction quality. We propose a spatiotemporal importance sampling approach to handle tool occlusion and large motion, further improving the rendering quality. Our experiments show that LerPlane achieves rendering quality comparable to or better than EndoNeRF in just three minutes, which is over 100x faster. Additionally, LerPlane enables information sharing across timesteps within the static field, reducing the impact of limited viewpoints. We believe our proposed method could inspire new pathways for robotic surgery scene understanding, empowering various downstream clinical-oriented tasks, and providing significant promise for future intraoperative surgery applications.

In this paper, we introduced LerPlane, a fast and accurate method for reconstructing deformable tissues from endoscopic videos. 
By utilizing multi-plane fields and spatiotemporal importance sampling, we can handle tool occlusion and large motion while significantly accelerating optimization. 
% LerPlane achieves comparable results with EndoNeRF in just $3$ minutes, over 100$\times$ faster.
Our experiments show that LerPlane achieves rendering quality comparable to or better than EndoNeRF in just three minutes, which is over 100 $\times$ faster.
We believe that LerPlane could improve robotic surgery scene understanding, benefiting various clinical-oriented tasks and intraoperative surgery applications.

In future work, our primary focus will be enhancing our approach's inference time to support intraoperative operations more efficiently. Additionally, we will dedicate efforts to reducing the requirements of input data, aiming to extend the applicability of LerPlane to a broader range of unconstrained surgical scenes.

\clearpage
{\small
\bibliographystyle{splncs04}
\bibliography{egbib}
}
\end{document}